Hate speech, Censorship, and Freedom of Speech:

The Changing Policies of Reddit

Elissa Nakajima Wickham
Waseda University
elissa@akane.waseda.jp

Emily Öhman
Waseda University
ohman@waseda.jp
ORC-iD: 0000-0003-1363-7361

Abstract

This paper examines the shift in focus on content policies and user attitudes on the social media platform Reddit. We do this by focusing on comments from general Reddit users from five posts made by admins (moderators) on updates to Reddit Content Policy. All five concern the nature of what kind of content is allowed to be posted on Reddit, and which measures will be taken against content that violates these policies. We use topic modeling to probe how the general discourse for Redditors has changed around limitations on content, and later, limitations on hate speech, or speech that incites violence against a particular group. We show that there is a clear shift in both the contents and the user attitudes that can be linked to contemporary societal upheaval as well as newly passed laws and regulations, and contribute to the wider discussion on hate speech moderation.

## I. Introduction

The peculiarities of online hate speech as compared to traditional, offline, hate speech has made it necessary to rapidly develop and implement laws regulating offensive speech online (Brown, 2018). The characteristics of online communities means that it is much easier for fringe ideals to gain foothold and members of such communities to spend their online lives in echo chambers of hate (Shaw, 2011), often targeting a specific group of people based on some real or perceived characteristic such as race, sex, sexuality, religion, and political beliefs.

The EU's 2016 Code on Countering Illegal Hate Speech Online (Code) introduced voluntary legislation that saw Twitter, Google, Facebook, and Microsoft commit to regulating hateful speech or speech that incites violence on their platforms. By 2020, each platform had stringent guidelines and flagging systems to remove reports of hate speech (Aswad, 2016). Anu Bradford (2019) would attribute this shift in practices to what she defined as "The Brussels Effect", or the regulatory power of the EU in certain markets to change standards of certain markets through their institutional capacity, willingness to create stringent rules, the relative market size of Europe, and the multilateral organization focusing on inelastic targets and non-divisible processes; meaning that their regulations had target aspects of the market that do not change, unlike capital or profit, and aspects that cannot be easily divided into separate processes to confront different standards of production.

In terms of the digital economy, Bradford (2019) identified two aspects where the Brussels Effect was in full force, specifically, the areas of data protection and hate speech. She was able to identify both the *de facto* and *de jure* aspect of the Brussels Effect in terms of data protection, as the EU had passed the 2016 binding legislation, the General Data Protection Regulation (GDPR), that forced a change in tech companies practices around the collection of user data. The *de facto* and *de jure* aspect of the Brussels Effect on hate speech remained unexplored, as the EU had not passed any binding legislation around hate speech regulation in the digital economy and only had the voluntary Code in effect at the time of writing her book; although currently, the binding version of the Code, The Digital Services Act (DSA) has been adopted and is in the process of being fully implemented at a later date.

Although the Code has been criticized by several scholars (see e.g. Aswad, 2016; Portaru, 2017; Alkiviadou, 2019) for both the privatization of hate speech making multinational private companies the arbiters of what constitutes legal and illegal hate speech and vague outcome reports (Portaru, 2017 - for the outcome report itself see Jourova, 2016), it has clearly facilitated change in how social media platforms deal with the datafication of hate (see e.g. Aswad, 2016; Laaksonen et al., 2020). As there is no universally accepted definition of hate speech (Nemes, 2010; Laaksonen, 2020) the private actors have little choice but to come up with their own definitions leading to vastly different policies on different social media platforms (Alkiviadou, 2019). Nonetheless, it is undeniable that the Code has greatly affected the content policies of all major social media platforms.

The overall effectiveness of the Code assists in understanding the EU's regulatory power in the realm of regulating hate speech online through the lens of the Brussels Effect. However, curiously enough, major American technology companies outside the scope of the Code have presented a distinct shift in philosophy surrounding their content policies and overall stance on hate speech since the adoption of the Code in 2016 (see e.g. Aswad, 2016; Bradford 2019). This offers the opportunity to make the argument that the EU's impact on hate speech practices is not only a result of their regulatory power, but a result of their normative power as well.

To highlight this phenomenon, it is only necessary to look at the Content Policy updates from the American online forum website, Reddit, to see how this shift has occurred. In 2015, Reddit expressed ambivalence towards the topic of hate speech, and rather demonstrated their commitment to libertarianism when it concerned violent speech. This is reflective of the general American political philosophy concerning hate speech, which ultimately enshrines offensive language in the First Amendment right to freedom of speech offered by the American constitution. In regards to offensive content, Reddit moderators state specifically, "It's ok to say 'I don't like this group of people.' It's not ok to say, 'I'm going to kill this group of people'" (Reddit - [u/spez], 2015a). They also cited their commitment to preventing "the speech police knocking down their door" (Reddit - [u/spez], 2015a). By 2020, Reddit's stance on hate speech would transform completely. They would state: "Everyone has a right to use Reddit free of harassment, bullying, and threats of violence. Communities and people that incite violence or that promote hate based on identity or vulnerability will be banned" (Reddit - [u/spez], 2020).

The change in Reddit Content policy represents a shift in practices from Reddit as a company, and this change necessitates the question of if general Reddit users have shifted their philosophy surrounding hate speech in line with the company. For this reason, this paper utilizes topic modeling of comments made by general Reddit users on Reddit admins' posts informing them of Content Policy updates to determine the general topics brought up in reaction to changing standards of what can and cannot be posted on the website. This is done in order to determine if and how the general philosophy of Reddit users has shifted on the topics on censorship and freedom of speech. We will first explain the context of the five content policy updates used for the analysis in this paper before presenting the data and methods used to conduct this analysis. We will then explore the results, before commenting on the change in the general lexicon of Reddit users and what it means for the future of social media platforms and further regulations of violent speech online.

## II.    Background

For a more comprehensive investigation of the reddit data we here qualitatively outline the content of each Reddit update utilized in the analysis for this paper. The Content Policy updates utilized in this paper begin in 2015, and end in 2020. The first update was made by the Reddit admin u/spez[1] in 2015 entitled, "Let's Talk Content. AMA[2]." in r/announcements[3] (Reddit - [u/spez], 2015a). This is the first time Reddit admins post about content restrictions on the website. An excerpt from this post reads:

> *As Reddit has grown, we've seen additional examples of how unfettered free speech can make Reddit a less enjoyable place to visit, and can even cause people harm outside of Reddit. Earlier this year, Reddit took a stand and banned non-consensual pornography. This was largely accepted by the community, and the world is a better place as a result (Google and Twitter have followed suit). Part of the reason this went over so well was because there was a very clear line of what was unacceptable.*
>
> Reddit - *[u/spez], 2015a*

---

[1] When referring to users on reddit, the username is traditionally preceded by a u/
[2] Ask Me Anything
[3] Communities on Reddit are known as subreddits and are denoted with an r/ preceding the community name.

This post displayed ambivalence towards hate speech, and focused more on content of a sexual nature, perhaps due to the rise on the #metoo movement which gained traction in 2013 and influenced several influential social media platforms' content policies (see e.g. Klonick, 2021). Twitter and Google, along with Facebook, had been under immense pressure at the time by German government officials to better their standards around the content posted on the site (see e.g. Alkiviadou, 2019). Notably, these companies would later become voluntary signatories to the EU's Code. This Reddit post was more of an open dialogue between users of Reddit and Reddit admins in preparation for their new Content Policy. Several researchers have pointed out the "laxness" of Reddit's content policies (Massanari, 2017; Gaudette et al., 2020). Incidentally, the #metoo movement and subsequent steps taken by social media platforms has had no impact on the prevalence of sexism in general (Archer et al., 2020), something that is echoed in many of the steps taken to combat hate speech online (Portaru, 2017).

The second post utilized for analysis in this paper is also from 2015 by Reddit admin u/spez in the r/announcements subreddit[4], where they introduced a "quarantine" function, which would essentially prevent subreddits that violated their new Content Policy from growing. An excerpt from the update read:

> *One new concept is Quarantining a community, which entails applying a set of restrictions to a community so its content will only be viewable to those who explicitly opt in. We will Quarantine communities whose content would be considered extremely offensive to the average redditor... Our most important policy over the last ten years has been to allow just about anything so long as it does not prevent others from enjoying Reddit for what it is: the best place online to have truly authentic conversations.*
>
> Reddit - *[u/spez], 2015b*

Again, it is possible to see the general ambivalence around the concept of hate speech. There is a focus on "offensive content", but at this point in time, Reddit admins mostly concern themselves with a general enjoyment of the website for users.The third post extracted for the analysis in this paper is from 2017, made by u/landoflobsters in r/modnews entitled, "Update on site-wide rules regarding violent content". It is important to note that this post was made one year after the introduction of the EU's Code. An excerpt from this post reads:

---

[4] Communities on Reddit are known as subreddits.

> *In particular, we found that the policy regarding "inciting" violence was too vague, and so we have made an effort to adjust it to be more clear and comprehensive. Going forward, we will take action against any content that encourages, glorifies, incites, or calls for violence or physical harm against an individual or a group of people; likewise, we will also take action against content that glorifies or encourages the abuse of animals. This applies to ALL content on Reddit, including memes, CSS/community styling, flair, subreddit names, and usernames.*

> Reddit - [u/landoflobsters], 2017

This post is the first time Reddit addresses violent speech, or in other words, hate speech. Specifically, they begin to update their restrictions around speech that incites violence against a particular individual or group of people.

The fourth post employed for analysis in this paper comes a year later, in 2018 with the title, "Revamping the quarantine function," made by u/landoflobsters in r/announcements. An excerpt from this post reads:

> *On a platform as open and diverse as Reddit, there will sometimes be communities that, while not prohibited by the Content Policy, average redditors may nevertheless find highly offensive or upsetting. In other cases, communities may be dedicated to promoting hoaxes (yes we used that word) that warrant additional scrutiny, as there are some things that are either verifiable or falsifiable and not seriously up for debate (eg, the Holocaust did happen and the number of people who died is well documented).*

> Reddit - *[u/landoflobsters], 2018*

While they do not mention hate speech in this update, the concept of removing hate speech can be identified in them announcing that subreddits dedicated to Holocaust denial, or other "hoaxes" are subject to a quarantine.

As for the effectiveness of quarantining Reddit communities, Chandrasekharan et al. (2021) examined two quarantined subreddits, The Red Pill[5] (r/theredpill, a misogynist community) and

---

[5] Red pills are a reference to the movie "The Matrix" (1999) where the protagonist is given a choice between a blue pill and a red pill. The red pill will make him see the world for what it is and the blue pill will allow him to keep his delusions. "Incels" believe that they have been given the proverbial red pill and can see women for "what they really are". They are male supremacist with extreme resentment towards women, often advocating sexual violence.

The Donald (r/the_donald, a racist community encouraging anti-Muslim content in particular). They found that although the influx of new users decreased significantly, despite the condition of exiting the quarantine being a significant reduction in sexist/misogynist and racist content, the offensive content levels remained at similar levels. Therefore quarantining worked to keep the hate from spreading, but did not alter the behavior of those who were already members of these communities nor did quarantining make them flee to other platforms (Chandrasekharan et al, 2021) or keep them from brigading (Gaudette et al, 2020). Brigading is  a practice where members of a specific Reddit community are encouraged to upvote specific, often offensive, content outside of the community itself to make it more visible to other users.

The final post utilized in this analysis announces Reddit's current Content Policy made by u/spez in r/announcements in 2020. An excerpt from this post states:

> *From our conversations with mods and outside experts, it's clear that while we've gotten better in some areas—like actioning violations at the community level, scaling enforcement efforts, measurably reducing hateful experiences like harassment year over year—we still have a long way to go to address the gaps in our policies and enforcement to date. These include addressing questions our policies have left unanswered (like whether hate speech is allowed or even protected on Reddit), aspects of our product and mod tools that are still too easy for individual bad actors to abuse (inboxes, chats, modmail), and areas where we can do better to partner with our mods and communities who want to combat the same hateful conduct we do.*

<div align="right">Reddit - [u/spez], 2020</div>

Here, Reddit clearly takes a stance on hate speech, and declares its commitment to the regulation and removal of hate speech on their platform. Four years after the adoption of the EU's Code, the general sentiment of Reddit admins concerning hate speech have demonstrated a clear shift in philosophy. The next step is to see how Reddit users have responded to these posts in the comments, and what these responses mean in terms of the philosophy of the general Reddit user.

III.    Data and Methods

Using the Reddit API Wrapper (PRAW[6]) for Python, all comments on each of the aforementioned five posts were scraped. The information we gained contained the username of each poster of the comment, the body of the comment made, and the upvote count of the comment.

Using structural topic modeling (STM) (Roberts et al., 2019) to gain insight into how much each covariant influences the topics of we focused on the 2015 AMA (Reddit - [u/spez], 2015a), the 2018 content policy update (Reddit - [u/landoflobsters], 2018), and the 2020 content policy update (Reddit - [u/spez], 2020). These three best reflect the evolution of content policy change on Reddit and gives us the best chance to objectively determine whether the attitude towards hate speech moderation and what should and should not be allowed in greater detail. We use the two additional posts for additional qualitative analysis.

For STM we determined that the best compromise between semantic coherence, exclusivity, and held-out likelihood measures was around 50 topics for the 2015 data, 27 for the 2018 data (see Figure 1), and 75 for the 2020 data. We used comment upvote scores as covariates. The STM package for R includes some built-in tools for pre-processing that include lower-casing, removal of punctuation, stopwords, and numbers, as well as stemming. Rare tokens are also removed. For example for the 2015 data there were originally 17988 terms of which 9364 were removed which is the equivalent of removing 9364 tokens out of 381422, i.e. tokens that appear only once in the data.

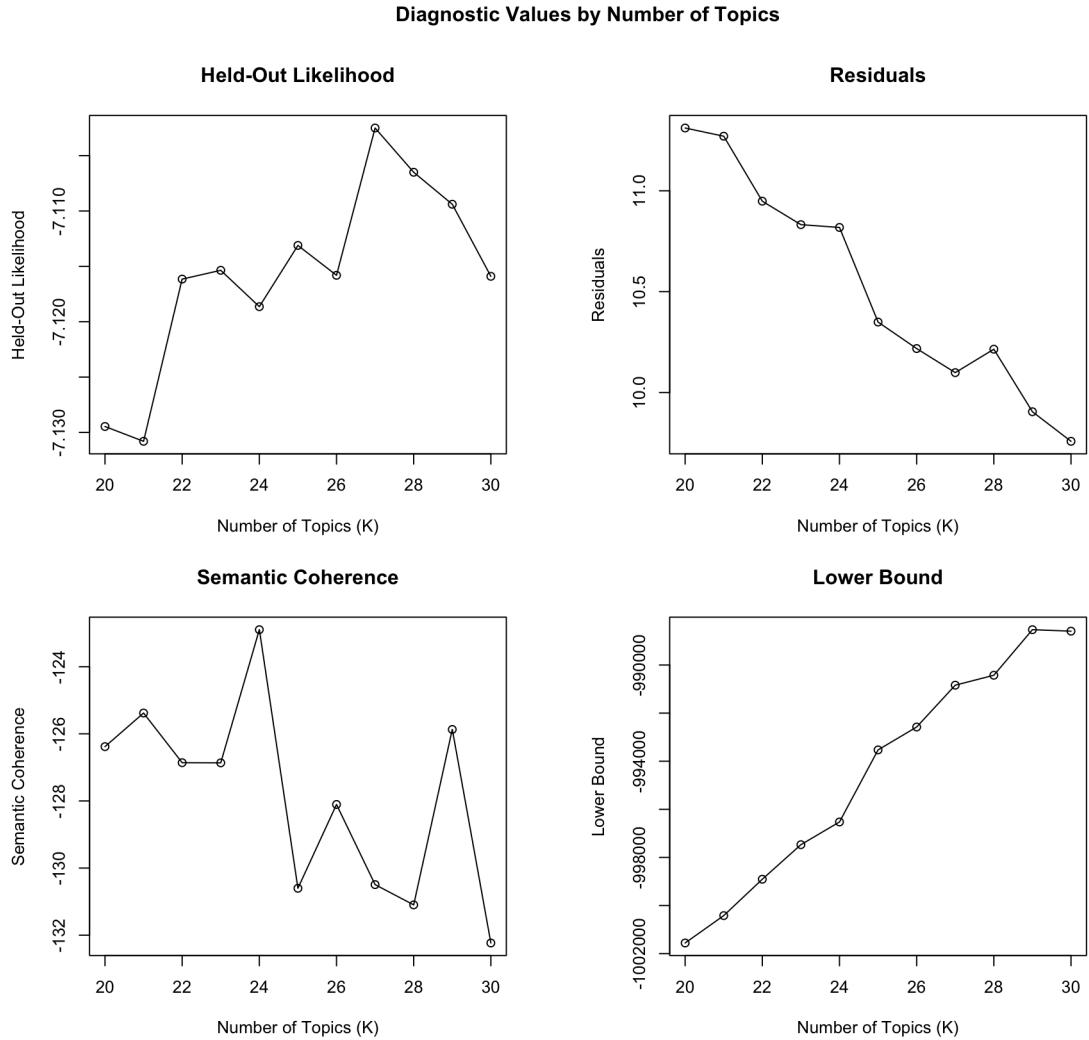

Figure 1 - Coherence measures for the Content Policy 2018 STM.

The three corpora examined using topic modeling after STM's pre-processing looked as follows:

| Corpus | # of documents[7] | # of terms | # of tokens |
|---|---|---|---|
| AMA 2015 | 18241 | 8624 | 372058 |
| Content 2018 | 7978 | 5580 | 130692 |
| Content 2020 | 32042 | 10298 | 472349 |

Table 1. Size of final corpora

---

[7] Documents here roughly correspond to individual comments, terms are unique tokens, and tokens the total number of tokens.

## IV.    Results

The most salient topics for each of the three STM are examined in this section. All topics and their FREX weighted[8] words can be viewed in the appendix. Here we only list the most representative topics and group them by common themes. We present these in chronological order starting with the initial AMA on Content Policy in 2015, continuing to the Content Policy update in 2018, and finishing with the Content Policy update in 2020.

The topics that emerge from the 2015 AMA (Reddit - [u/spez], 2015) revolve around the need for more transparency around bans and rules (see topics 5, 18, 30, 31, 34, 40, 48), the need for rules around brigading subreddits (see topics 1, 13, 14, 24, 25, 40), and separating valid criticism from real threat (see topics 4, 16, 21, 47 plus doxxing: topic 9). There is also a secondary discussion going on that deals with censorship in general (topics 4, 8, 10, 11, 27) and Reddit as a corporation and how potential revenue influences decisions on content policy (topic 36, 39, 45).

Topic 24 Top Words:
 Highest Prob: harass, group, anyth, peopl, bulli, individu, other
 FREX: intimid, bulli, silenc, behavior, individu, group, harass
 Lift:<URL>,<URL>, <URL>, oklet, intimid, hive-mind
 Score: harass, bulli, group, intimid, individu, silenc, abus
Topic 27 Top Words:
 Highest Prob: speech, free, freedom, express, protect, allow, say
 FREX: speech, freedom, free, express, consequ, unfett, principl
 Lift: farewel, fetter, <URL>, <URL>, <URL>, triggeredteehe
 Score: speech, free, freedom, express, protect, unfett, consequ
Topic 30 Top Words:
 Highest Prob: subreddit, rule, allow, encourag, enforc, mani, break
 FREX: rule, subreddit, encourag, break, enforc, warn, guidelin
 Lift: banana, haikus, hiaku, ponzi, removedban, throwingpotatosatclown, doot
 Score: subreddit, rule, encourag, enforc, break, allow, guidelin
Topic 34 Top Words:
 Highest Prob: will, content, offens, sens, violat, see, defin
 FREX: violat, decenc, content, sens, will, list, common
 Lift: silo, decenc, login, usefulbot, checkbox, futhermor, advisori
 Score: content, will, decenc, violat, sens, nsfw, common

---

[8] FREX shows how common AND exclusive a word is for a particular topic, i.e. it is the highest probability words weighted for exclusivity. Lift is calculated by dividing the topic-word distribution by the empirical word count probability distribution. Score is calculated as $\beta_{v, k} (\log \beta_{w,k} - 1 / K \sum_{k'} \log \beta_{v,k'})$ where $\log \beta$ is a K by V matrix containing the log probabilities of seeing word v conditional on topic k (Roberts et al., 2019).

**Top Topics**

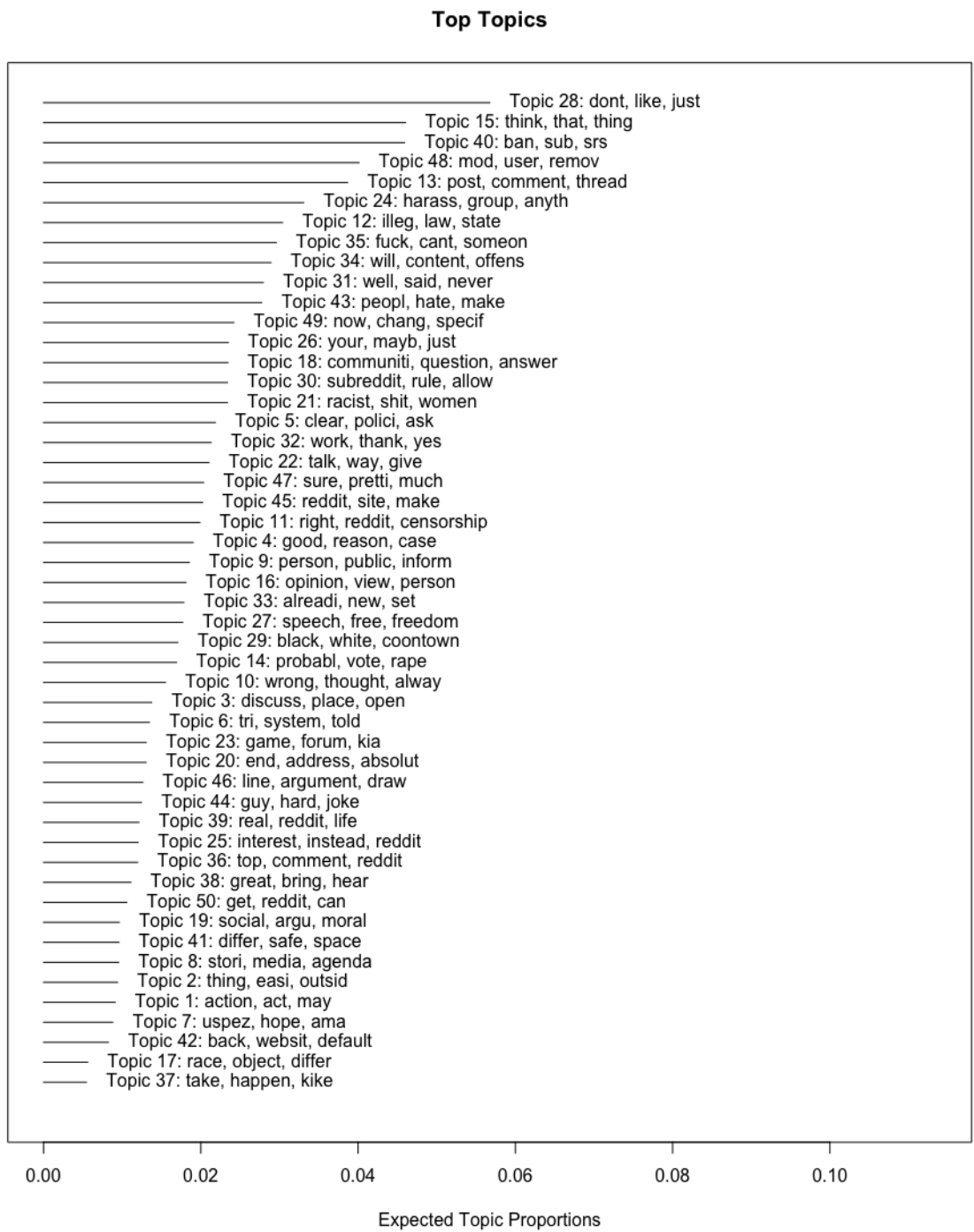

Figure 2. Expected topic proportions for AMA 2015.

The 2018 update's most salient topics revolve around the quarantine function, and general Reddit users backlash against what is perceived as censorship. Specific topics are free speech and censorship (see topics 1, 3, 7, 15, 20), banning violent content and propaganda (topics 4, 14), as well as the logistics of quarantined subreddits (topics 23). The conversation on Reddit as a corporation is also still active (topics 8, 9) as is the issue with brigading (topic 10). There are also many references to politics and political actors (topics 4, 5, 6, 12, 19) and racism (topics 15, 16, 17, 22). A meta discussion relating to both free speech and censorship as well as propaganda can be inferred from topic 20.

Topic 3 Top Words:
> *Highest Prob*: ban, speech, hate, free, read, first, exact
> *FREX*: voat, hate, echochamb, ban, hill, speech, infest
> *Lift*: brightest, durr, lib, alan, articul, banning", blackpeopletwitt
> *Score*: ban, speech, hate, free, read, voat, first

Topic 23 Top Words:
> *Highest Prob*: sub, quarantin, can, communiti, user, content, list
> *FREX*: quarantin, nsfw, accident, revenu, sub, heads-, search
> *Lift*: "ban", aboutjson, accident, accross, blank, canari, commentspost
> *Score*: quarantin, sub, content, user, communiti, offens, list

**Top Topics**

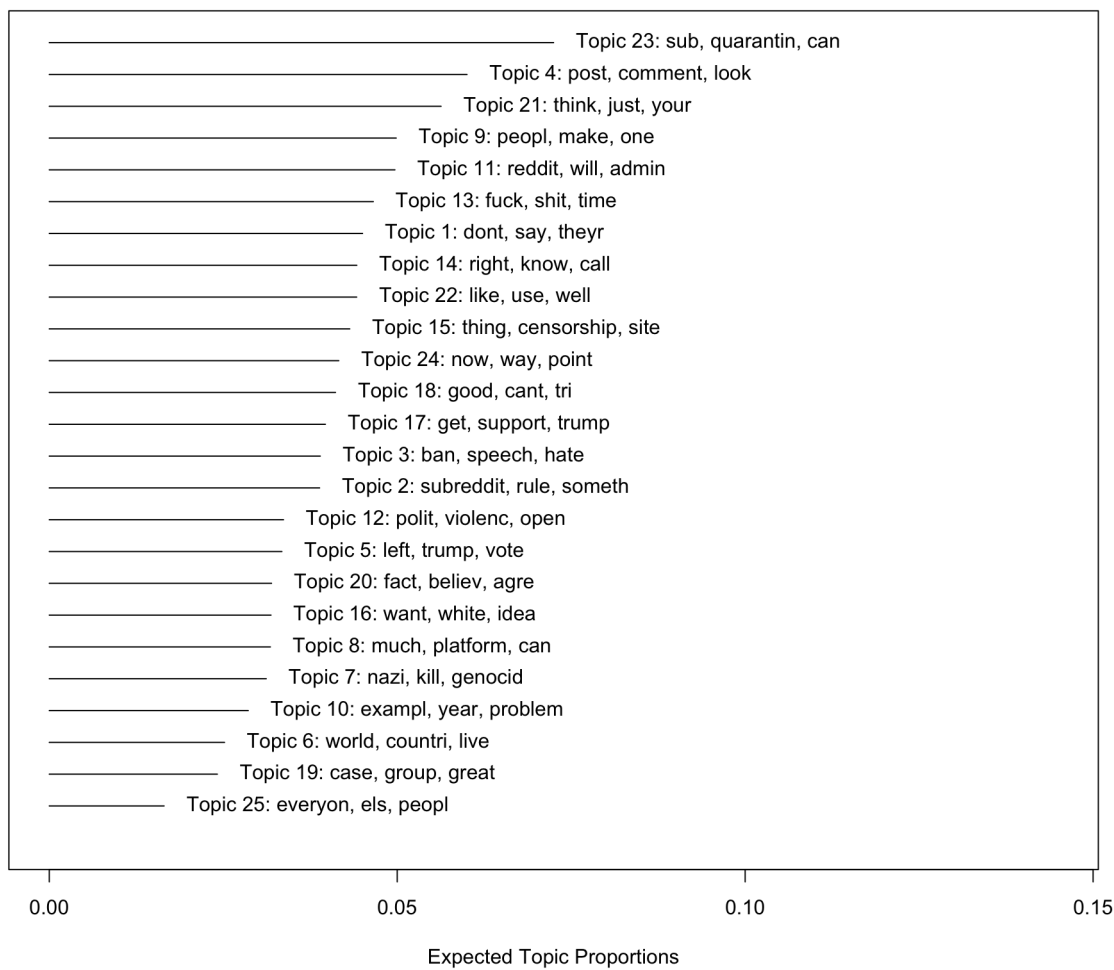

Figure 3. Content Policy 2018 post topic proportions

The 2020 Content Policy update's most salient topics express the collective desire of Redditors to create safety for marginalized groups (topics 11, 17, 19, 29, 32, 34, 35, **37**, 40) taking into account the power of Reddit as a tool for discrimination (topics 2, 16, 17, 22, 39, 63). There is also very specific talk around racism and to a lesser extent sexism (topics 46, 48, 68, 74) and rule enforcement (topic 27). Some old topics re-emerge like the discussion on banning subreddits (12, 15, 52) as well as censorship in general (36, 61). A new but very prevalent topic is that of Chinese propaganda on the platform, often specifically regarding the subreddit r/Sino (topics 5, 6, 7, 50).

Topic 5 Top Words:
  *Highest Prob*: great, squar, tiananmen, forward, 李洪志, 法輪功, tibet
  *FREX*: squar, tiananmen, 李洪志, 法輪功, tibet, 天安門, 天安门
  *Lift*: tibet, abduct, anti-protect, anti-rightist, anti-riot, baggag, epoch
  *Score*: 李洪志, 法輪功, 天安門, 天安门, great, falun, tiananmen
Topic 7 Top Words:
  *Highest Prob*: china, countri, system, parti, cultur, america, democrat
  *FREX*: ccp, cancel, own, democraci, tencent, parti, econom
  *Lift*: conspiratori, militar, anti-asian, asset, ccp, cpc, denmark
  *Score*: china, democrat, system, countri, america, cultur, parti
Topic 37 Top Words:
  *Highest Prob*: major, group, protect, rule, hate, minor, promot
  *FREX*: group, minor, major, protect, promot, global, margin
  *Lift*: "disenfranchised", "hangfs", "killallnrs", "major, "protect, disability", hate"
  *Score*: group, major, protect, rule, ident, hate, minor

**Top Topics**

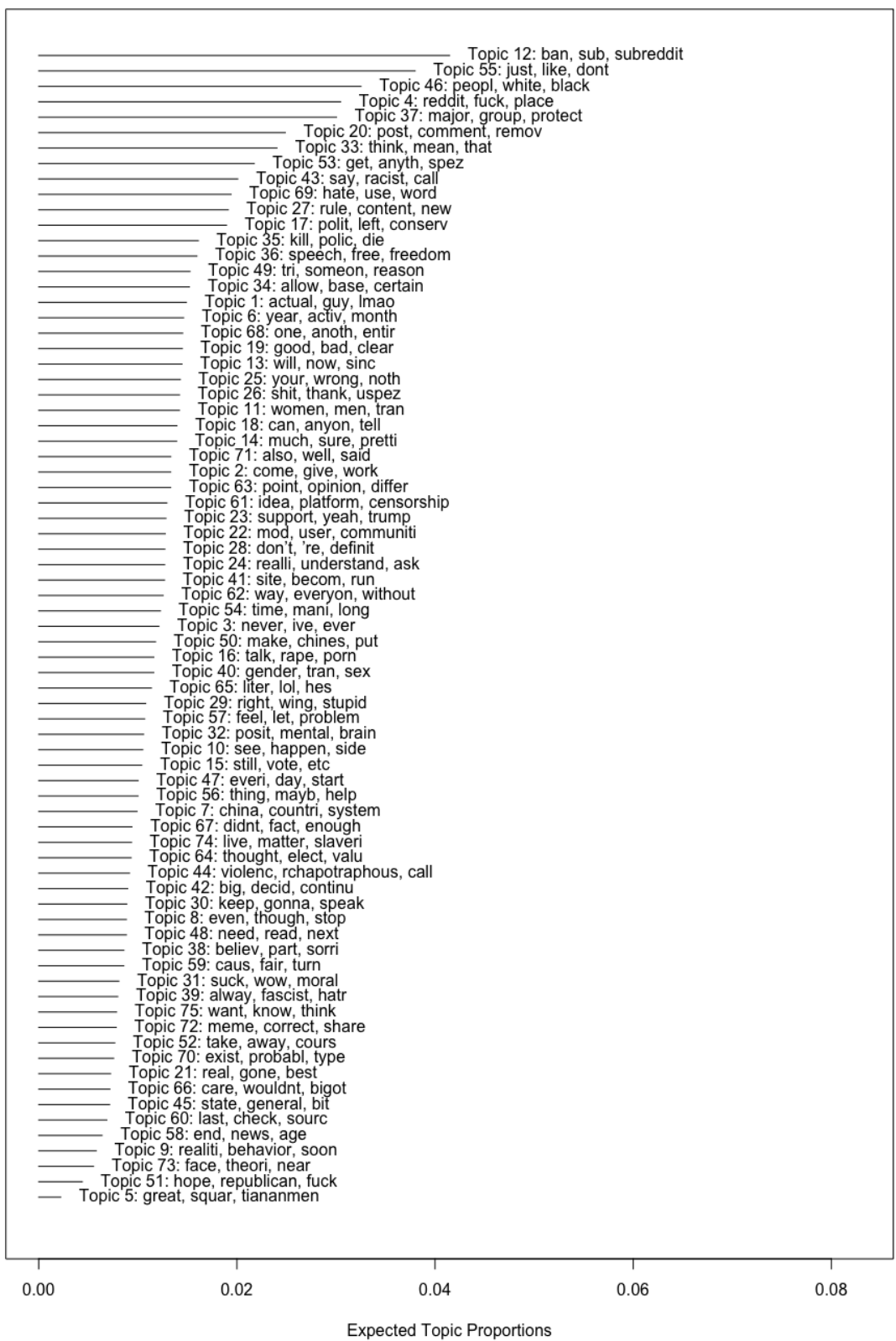

Figure 4. Expected topic proportions of the 2020 Content Policy update.

## V.    Analysis

The top salient topics from both Content Policy posts in 2015 reflect an openness of Reddit users to rules around banning certain kinds of topics or subreddits dedicated to "brigading", a form of harassment on the website where users from one subreddit enter another to flood the comment and post section with harassment or other irrelevant content. The general openness of users to more rules would support the assertion Anu Bradford (2019) makes about hate speech regulation and the EU in the digital economy; that the EU has the strictest rules that the platforms respond to, unlike other regulators like Russia and China that also have strict rules that seem to repulse the most popular platforms. One year after the major signatories adopted the EU's Code, Reddit would also find themselves clarifying what violent speech meant in terms of acceptable content on the site.

A look at the update on rules around violent content in 2017 suggests that when the topic of violent speech is introduced by the admins, general users are able to engage with this topic, and provide examples of a real problem within the Reddit community concerning violent speech; in this case, violent speech directed towards women by incels, who blame women and society for their inability to find a sexual partner. Violent speech would remain a problem that necessitated a solution, without much controversy, although the regulation of some other types of hate speech showed some resistance. It seems as though the "incel problem" was less controversial in 2017, as incels were fringe private citizens with generally distasteful views. This would change, however, for general Redditors once politics were introduced into the discussion.

The most notable shift is from the top salient topics in the post regarding quarantines in 2018, to the top salient topics in the post announcing Reddit's current hate speech policy in 2020. 48.93% of  Reddit users are American (Clement, 2021), and 2018 marked the second year of 45th President Donald Trump's time in office. The post from 2018 remarked upon "hoax' subreddits, and directly mentioned subreddits dedicated to Holocaust denial (topic 7). At that time, r/The_Donald was a controversial subreddit at the time that was dedicated to Donald Trump and fostered far-right discourse and conspiracy theories (topics 5, 17). Holocaust denial is only one of many conspiracies parroted by the far-right in America (Souther Poverty Law Center, 2021). It would later be quarantined in 2019 (Haskins, 2019), then fully banned in 2020. A Reddit spokesperson told *Vice* news that 'we are sensitive to what could be considered political speech, however, recent behaviors including threats against the police and public figures is content that is prohibited by our violence policy. As a result, we have… quarantined the subreddit" (Haskins, 2019). The backlash against the initial post about revamping the quarantine system and applying them more liberally can be understood as an initial reaction against the perceived suppression of political views. With the majority of Reddit users being American, this would reflect a distaste for anything that could potentially violate the concept of freedom of speech in the First Amendment of the American Constitution. Ultimately, the subreddit dedicated to Donald Trump, would, in fact, stay unquarantined for a full year before finally receiving consequences.

In the 2018 update there is an ongoing discussion where a minority, but a significant minority,  of Reddit users seem willing to ban violent speech even if it meant censorship. This minority would become the majority by 2020. It was not until 2020, when the US experienced the largest civil rights movement in its history, Black Lives Matter, a movement against the asymmetrical treatment and killings of black Americans by the police, that Reddit moved to protect groups of people in their formal content policy (visible in topics 43 & 46 in particular). At this point in time, Twitter had begun to flag then-President Donald Trump's tweets for incitement of violence as he responded to the civil rights movement with a call to retaliatory violence against protestors who had looted and rioted in certain areas (Twitter, 2021). It is arguable that at this point, social media platforms themselves began to label the philosophy of Donald Trump as intrinsically violent. In 2018, users may not have agreed, but by 2020, as shown by the topics brought up by Reddit users, they too had come to the same conclusion.

## VI.    Discussion

How the EU has been able to impact normative behavior surrounding hate speech online is insufficiently explained through the Brussels Effect in the digital economy. While Bradford (2019) makes a compelling argument for the regulatory power of the EU in the online realm, the changes observed in Reddit moderation and admin teams' attitude toward regulation of offensive content offer a deeper, perhaps unintentional consequence of the interaction of differing political actors in the internet realm. The US Supreme Court views hate speech as a fundamental aspect of the democratic right to freedom of speech. In 2017, the US Supreme Court ruled on *Matal vs.*

*Tam*, a case concerning the trademark registration of the band name, "The Slants"; stating that the "reclaiming" of the derogatory term for Asians was protected under free speech:

> *The Government has an interest in preventing speech expressing ideas that offend. And, as we have explained, that idea strikes at the heart of the First Amendment. Speech that demeans on the basis of race, ethnicity, gender, religion, age, disability, or any other similar ground is hateful; but the proudest boast of our free speech jurisprudence is that we protect the freedom to express "the thought that we hate."*

Matal vs. Tam, 2017

The freedom to express hateful opinions is part of a larger American philosophy that the US Supreme Court previously discussed in the 1974 case of *Miami Herald Publishing Co. v. Tornillo,* with the ruling that "the widest possible dissemination of information from diverse and antagonistic sources is essential to the welfare of the public".

The philosophy of Reddit's aforementioned content policy update in 2015 mirrors the general sentiments of the Court on the topic of hate speech. The Court refers to this philosophy as the "freedom to express 'the thought that we hate'", while Reddit states that "people have more open and authentic discussions when they aren't worried about the speech police knocking down their door" (Reddit, 2015a). This philosophy would also be mirrored in the discourse of general Reddit users in the replies to this update; topic three highlights this concern in the highest probability terms and FREX terms: "discuss, place, open, reddit, plan, polic(y/ies), restrict / authent, conflict, worri, plan, discuss, door, restrict'

The EU takes an opposite approach to hate speech, viewing violent language as a threat to democratic freedom as a whole, as evidenced by the ECHR's ruling in *Refah Partisi (The Welfare Party) and Others v. Turkey*, where it was asserted that it was allowable for there to exist "some compromise between the requirements of defending democratic society and individual rights" (2003). Reddit's most recent content policy update in 2020 would reflect the European philosophy of compromise as opposed to the American philosophy of unregulated discussion: "it's our responsibility to support our communities by taking stronger action against those who try to weaponize parts of Reddit against other people" (Reddit, 2020).

The reason why Reddit is notable in the discussion of the EU's ability to impact norms in the online sphere is because Reddit is currently outside the scope of EU legislation, binding and non-binding. The differing philosophies of the US and EU on hate speech centers around the meaning of individual rights in the context of a free and democratic society. Thus, this conflict can be understood as a tension around the meaning of human rights between the EU and the US; with the EU's introduction of hate speech regulation online as a distinct effort to influence human rights norms in American tech companies who operate in an environment that holds contrary opinions.

Risse and Sikkink (1999) would identify the processes through which human rights norms are ultimately progressed from "commitment to compliance". Because political actors ultimately rely upon the recognition from, and interaction with, other political actors; it then becomes possible for different norms of human rights to be socialized into other regions. This process of socialization embodies three separate processes. The first process is "instrumental adaptation", where actors, upon mounting pressure from other actors, make "tactical concessions" in the human rights realm, whether that involves making formal commitments or signing binding treaties. The second process involves "argumentative discourses" that involve, "shaming and denunciations, not aimed at producing changing minds with logic, but on changing minds by isolating or embarrassing the target" (Risse and Sikkink, 1999). The third and final process concerns "institutionalization and habitualization", in which the norms being diffused are viewed by the actor as the "normal thing to do".

The evolution of Reddit's content policy is evidence of the third process of the EU's concentrated effort to socialize American tech platforms to its own conception of human rights and hate speech, to remarkable success. While Reddit and other tech platforms are not distinct political actors as originally envisioned by Risse and Sikkink (1999), it is possible to argue that the two did not anticipate the prevalence of social media in daily life in 1999, and today, several arguments have been made for tech platforms to be considered as distinct political entities (Helberger, 2020; Gilardi, 2021).

In 2015, the EU would see the first processes of argumentation, specifically, naming and shaming when German Justice Minister Heiko Maas threatened Facebook with regulations on a parliamentary level if the company neglected to address the growing prevalence of hate speech on the platform (DW, 2015). Instrumental adaptation would begin for American tech platforms in the dialogue that began between the EU and US social media sites following Maas' public callout, that eventually birthed the 2016 Code on Countering Illegal Hate Speech Online (Oltermann, 2016) that made the removal of hate speech standard practice. It can easily be argued that the EU was effective in socializing platforms to hate speech norms, as nearly every major American tech platform holds some level of restriction on offensive language. However, scrutiny of the effectiveness of the Code in socializing *users* of these platforms to European norms around hate speech is difficult; as platforms like Facebook, Twitter, and Google do not offer the opportunity for users to directly interact with changing content policy. With the option for users to directly interact with changing policy, Reddit creates the unique opportunity to examine the shifting dialogue within and between users as hate speech removal becomes standard in American tech companies.

When considering the prevalence of an expressed need to protect marginalized groups in the top salient topics from 2020; while the debate in general may not fully reflect a complete commitment to European standards over American ones, it is clear that the ideas that motivated the EU's Code in the first place have entered the consciousness of internet users. In other words, the standards of hate speech removal on platforms, as first enshrined in the Code, has been accepted by a good portion of users as normal; offering not only evidence for the EU's regulatory power in this realm, but its power in shaping the norms of the internet to their conception of individual rights. After the banning of Donald Trump and related accounts by Reddit and several other platforms, a spokesperson for Angela Merkel, chancellor of Germany at the time, released this statement in response: "...the fundamental right [of freedom of expression] can be interfered with, but along the lines of the law and within the framework defined by the lawmakers. Not according to the decision of the management of social media platforms" (Bermingham, 2021). While the debate continues surrounding which actor has the right to remove content from platforms; Merkel's comment and the results of this analysis make one thing clear: hate speech removal is now standard practice in the minds of moderators and policymakers and in the minds of users as well.

## VII. Future Work

Scraping general user data and topic modeling what they say about content policy updates and the freedom of speech and censorship can be extended to other social media platforms under the scope of the EU, as well as outside the scope of the EU, as has been done here. Furthermore, sentiment analysis could be utilized to not only discover what topics are being brought up by users in response to hate speech regulation, but also to analyze the sentiments attached to each topic.

Future work would include sentiment analysis of the body of the comments on each Reddit post and using these sentiments over time as covariates in the STM. We would also like to explore statements, their topics and sentiments, at different points in time on other social media platforms.

# Appendix

Topic list for 2015 AMA:

---

Topic 1 Top Words:
     Highest Prob: action, act, may, particular, term, form, equal
     FREX: action, associ, justifi, act, sentenc, equal, thus
     Lift: calappd, calappth, calrptr, calrptrd, courtroom, hankin, omit
     Score: action, act, associ, justifi, discrimin, form, equal
Topic 2 Top Words:
     Highest Prob: thing, easi, outsid, busi, fit, main, horribl
     FREX: easi, anonym, horribl, fit, stick, togeth, replac
     Lift: --thinly-veil, <URL>, searchal, sibl, uchesterhiggenbothum, woosh, ◢ ⌐∩⌐
     Score: busi, easi, horribl, anonym, section, fit, main
Topic 3 Top Words:
     Highest Prob: discuss, place, open, reddit, plan, polic, restrict
     FREX: authent, conflict, worri, plan, discuss, door, restrict
     Lift: ayyyi, pages—, reddit—, vocabulari, authent, site", prolong
     Score: discuss, open, authent, plan, restrict, conflict, worri
Topic 4 Top Words:
     Highest Prob: good, reason, case, subject, decis, standard, need
     FREX: court, good, standard, suprem, decis, lack, subject
     Lift: prurient, concur, excretori, <URL>, <URL>, literari, miller
     Score: good, court, standard, case, subject, decis, reason
Topic 5 Top Words:
     Highest Prob: clear, polici, ask, seem, respons, exact, sinc
     FREX: polici, ask, clear, friend, respons, offici, sort
     Lift: <URL>, <URL>, <URL>, <URL>, <URL>, piggyback, ofpeopl
     Score: polici, ask, clear, respons, friend, sort, intend
Topic 6 Top Words:
     Highest Prob: tri, system, told, can, live, sometim, time
     FREX: tri, told, reflect, system, sometim, unfortun, convinc
     Lift: chucklefuck, gamemast, <URL>, lobbingpotatoesatclown, pnp, rck, rome
     Score: tri, system, told, sometim, unfortun, reflect, period
Topic 7 Top Words:
     Highest Prob: uspez, hope, ama, surpris, gonna, worst, drama
     FREX: ama, uspez, advic, shitstorm, yishan, gonna, surpris
     Lift: pre-prepar, rblog, spaz, theyeticaptain, umdri, uusersimul, <URL>
     Score: uspez, ama, hope, gonna, surpris, worst, afraid
Topic 8 Top Words:
     Highest Prob: stori, media, agenda, push, chan, toxic, leak
     FREX: reurop, riot, stori, agenda, chan, kotakuinact, leak
     Lift: fondl, gerstmann, relent, reurop, rioter, tendril, tenuous
     Score: reurop, agenda, media, stori, chan, gamerg, leak
Topic 9 Top Words:
     Highest Prob: person, public, inform, name, pictur, post, doxx
     FREX: photo, pictur, name, imgur, confidenti, inform, pic
     Lift: astronaut, cereal, collag, gaf, isntt, slimgur, super-leet
     Score: name, public, pictur, imgur, inform, doxx, photo
Topic 10 Top Words:
     Highest Prob: wrong, thought, alway, mind, side, human, correct
     FREX: thought, alway, happi, board, wrong, correct, dark

Lift: shade, sheepl, highi, <URL>, <URL>, <URL>, hubri
Score: thought, alway, wrong, side, correct, happi, human

Topic 11 Top Words:
Highest Prob: right, reddit, censorship, censor, privat, speak, govern
FREX: bastion, right, speak, govern, censorship, amend, censor
Lift: ayyyyy, <URL>, inalien, ngger, spraypaint, speach, unworthi
Score: right, bastion, govern, censorship, privat, censor, amend

Topic 12 Top Words:
Highest Prob: illeg, law, state, link, copyright, host, reddit
FREX: copyright, illeg, weed, dmca, law, download, drug
Lift: buysel, cali, cloudflar, contributori, dmca, fatsham, feloni
Score: illeg, copyright, law, materi, drug, host, legal

Topic 13 Top Words:
Highest Prob: post, comment, thread, look, read, time, downvot
FREX: downvot, thread, read, post, upvot, comment, respond
Lift: lake, mach, scoop, <URL>, rredditaltern, shore, topthi
Score: post, comment, thread, downvot, upvot, read, look

Topic 14 Top Words:
Highest Prob: probabl, vote, rape, subscrib, get, less, might
FREX: karma, vote, probabl, nah, subscrib, rape, solut
Lift: cannib, certif, font, fuzz, gilberto, <URL>, irreconcil
Score: vote, rape, probabl, karma, subscrib, solut, altern

Topic 15 Top Words:
Highest Prob: think, that, thing, theyr, doesnt, problem, everyon
FREX: that, theyr, think, everyon, better, agre, problem
Lift: compound, <URL>, litig, flippiti, hah, usupcaci, that
Score: that, think, thing, theyr, better, doesnt, everyon

Topic 16 Top Words:
Highest Prob: opinion, view, person, polit, attack, feel, disagre
FREX: echo, opinion, chamber, oppos, attack, view, viewpoint
Lift: 'grey, 'safe, 'sensitivity', "content, anti-porn, area', content-neutr
Score: opinion, view, attack, polit, chamber, echo, viewpoint

Topic 17 Top Words:
Highest Prob: race, object, differ, one, use, measur, valu
FREX: basketbal, object, measur, player, notion, cluster, race
Lift: analys, ascertain, classificatori, heurist, non-sickle-cel, pan, solar
Score: race, biodivers, subspeci, haplogroup, basketbal, object, genet

Topic 18 Top Words:
Highest Prob: communiti, question, answer, reddit, tool, can, issu
FREX: communiti, answer, question, process, communic, forward, tool
Lift: decentr, patch, spook, akron, doucher, gasolin, honorari
Score: communiti, answer, question, tool, communic, process, reddit

Topic 19 Top Words:
Highest Prob: social, argu, moral, respect, justic, ration, challeng
FREX: moral, social, respect, warrior, challeng, rude, killer
Lift: beginn, india, medicin, nudg, vitamin, <URL>, fempiresrd
Score: moral, social, respect, justic, argu, ration, killer

Topic 20 Top Words:
Highest Prob: end, address, absolut, nobodi, one, easili, get
FREX: christ, sick, absolut, end, jesus, rcandidfashionpolic, address
Lift: apathet, ipv, nazareth, putrid, telephon, tract, veneer
Score: end, address, jesus, absolut, somebodi, christ, nobodi

Topic 21 Top Words:
Highest Prob: racist, shit, women, serious, men, lol, sjw
FREX: men, shit, trp, lol, women, sjw, hilari
Lift: -colour, chromosom, fedora, gasthekek, gsm, non-smok, pussypass

Score: racist, shit, women, men, lol, sjw, serious

Topic 22 Top Words:

Highest Prob: talk, way, give, let, anyon, know, wouldnt

FREX: anyon, give, talk, wouldnt, let, bother, doubt

Lift: ceown, <URL>, <URL>, <URL>, <URL>, <URL>, <URL>

Score: anyon, give, hodor, talk, wouldnt, let, way

Topic 23 Top Words:

Highest Prob: game, forum, kia, level, play, video, miss

FREX: warlizard, kia, game, journal, holi, level, play

Lift: errand, gearbox, jabroni, pray, rfirespez, simciti, "master

Score: game, kia, journal, forum, ethic, level, warlizard

Topic 24 Top Words:

Highest Prob: harass, group, anyth, peopl, bulli, individu, other

FREX: intimid, bulli, silenc, behavior, individu, group, harass

Lift: <URL>, <URL>, <URL>, <URL>, oklet, intimid, hive-mind

Score: harass, bulli, group, intimid, individu, silenc, abus

Topic 25 Top Words:

Highest Prob: interest, instead, reddit, decid, topic, popular, hand

FREX: popular, instead, topic, interest, select, hand, parent

Lift: eta, illuminati, super-downvot, tryhard, antithesi, lectur, demmian

Score: topic, interest, instead, popular, hand, decid, news

Topic 26 Top Words:

Highest Prob: your, mayb, just, offend, love, bullshit, okay

FREX: okay, your, mayb, asshol, love, idiot, bullshit

Lift: disk, reptil, sopa, stimuli, unhing, clifford, beimg

Score: your, mayb, bullshit, love, idiot, asshol, offend

Topic 27 Top Words:

Highest Prob: speech, free, freedom, express, protect, allow, say

FREX: speech, freedom, free, express, consequ, unfett, principl

Lift: farewel, fetter, <URL>, <URL>, <URL>, <URL>, triggeredteehe

Score: speech, free, freedom, express, protect, unfett, consequ

Topic 28 Top Words:

Highest Prob: dont, like, just, want, realli, see, say

FREX: dont, realli, want, like, care, ill, see

Lift: hatebrigad, nato, shhhhhhhh, uchristaliaferro, brazilian, discord, utc

Score: dont, like, want, just, realli, see, say

Topic 29 Top Words:

Highest Prob: black, white, coontown, kill, rcoontown, man, die

FREX: nigger, coontown, die, black, commit, white, roof

Lift: plastic, cya, flush, haunt, hippi, <URL>, munch

Score: black, white, coontown, kill, crime, nigger, supremacist

Topic 30 Top Words:

Highest Prob: subreddit, rule, allow, encourag, enforc, mani, break

FREX: rule, subreddit, encourag, break, enforc, warn, guidelin

Lift: banana, haikus, hiaku, ponzi, removedban, throwingpotatosatclown, doot

Score: subreddit, rule, encourag, enforc, break, allow, guidelin

Topic 31 Top Words:

Highest Prob: well, said, never, didnt, use, shadowban, guess

FREX: never, well, spez, shadowban, shadow, said, didnt

Lift: <URL>, <URL>, <URL>, <URL>, <URL>, <URL>, title-text

Score: well, said, shadowban, never, didnt, nice, shadow

Topic 32 Top Words:

Highest Prob: work, thank, yes, keep, wont, edit, yeah

FREX: thank, donger, bot, wont, edit, yeah, check

Lift: blowjob, toast, ヽ༼ຈل͜ຈ༽ﾉ, donger, dork, entryism, musician

Score: thank, yeah, yes, wont, bot, edit, spam

Topic 33 Top Words:
        Highest Prob: alreadi, new, set, sourc, tag, like, add
        FREX: tag, rall, sourc, nsfl, alreadi, overwritten, \<URL>
        Lift: ofnv, hintus, possib, \<URL>, useraposs, overwritten
        Score: tag, nsfw, sourc,\<URL>, useraposs, alreadi, nsfl
Topic 34 Top Words:
        Highest Prob: will, content, offens, sens, violat, see, defin
        FREX: violat, decenc, content, sens, will, list, common
        Lift: silo, decenc, login, usefulbot, checkbox, futhermor, advisori
        Score: content, will, decenc, violat, sens, nsfw, common
Topic 35 Top Words:
        Highest Prob: fuck, cant, someon, call, tell, arent, stop
        FREX: fuck, cant, stupid, sorri, tell, anymor, liter
        Lift: outta, ping, sbinp, setuid, existenti, ghostfac, liveleak
        Score: fuck, cant, stupid, liter, tell, call, arent
Topic 36 Top Words:
        Highest Prob: top, comment, reddit, pao, ceo, click, certain
        FREX: \<URL>, ceo, pao, join, due, ellen, top
        Lift: \<URL>, lactos, preferenti, trumped-, unjustifi, updoot, \<URL>
        Score: overwrit, \<URL>, \<URL>, greasemonkey, \<URL>-monkey, \<URL>, safari
Topic 37 Top Words:
        Highest Prob: take, happen, kike, overthi, nutshel, overv, rmurica
        FREX: kike, happen, overthi, take, overv, nutshel, rmurica
        Lift: kike, overthi, overv, happen, take, nutshel, rmurica
        Score: overthi, kike, take, happen, overv, nutshel, rmurica
Topic 38 Top Words:
        Highest Prob: great, bring, hear, intent, threaten, person, organ
        FREX: threaten, hear, great, bring, repetit, disturb, touch
        Lift: pester, thankyou, washington, big-budget, derek, duck-siz, horse-s
        Score: threaten, great, hear, bring, articl, organ, intent
Topic 39 Top Words:
        Highest Prob: real, reddit, life, lose, move, corpor, internet
        FREX: holocaust, denial, digg, lose, corpor, traffic, powermod
        Lift: atroc, bbss, monetari, fetus, latent, mattress, rbadcopnodonut
        Score: real, corpor, lose, holocaust, digg, traffic, life
Topic 40 Top Words:
        Highest Prob: ban, sub, srs, brigad, fph, link, harass
        FREX: brigad, srd, srs, ban, fph, proof, evid
        Lift: -someth, gasthekik, nineti, askmen, \<URL>, evas, \<URL>
        Score: ban, sub, fph, srs, brigad, link, harass
Topic 41 Top Words:
        Highest Prob: differ, safe, space, terribl, describ, race, one
        FREX: safe, space, differ, terribl, amaz, describ, scare
        Lift: down, reedit, arisen, biologist, breastplat, combust, either-
        Score: differ, safe, space, race, subspeci, terribl, speci
Topic 42 Top Words:
        Highest Prob: back, websit, default, hate, time, reddit, recruit
        FREX: back, largest, chimpir, reput, ralli, fester, blah
        Lift: antebellum, captain, crank, litter, paosuekjp, peel, phrenolog
        Score: back, websit, default, rvideo, blah, hate, chimpir
Topic 43 Top Words:
        Highest Prob: peopl, hate, make, fat, fun, hurt, suicid
        FREX: fat, hate, fun, suicid, shame, overweight, rfatlog
        Lift: lulz, glutton, \<URL>, mow, nutrit, pothead, rfeministfrequ
        Score: peopl, hate, fat, fun, suicid, rfatpeopleh, fatpeopleh
Topic 44 Top Words:

Highest Prob: guy, hard, joke, total, get, hey, cool
FREX: guy, rekt, cool, hey, joke, hard, ton
Lift: <URL>, lighten, rekt, grape, neon, parliament, shrekt
Score: guy, rekt, joke, cool, hard, hey, sjws

Topic 45 Top Words:
Highest Prob: reddit, site, make, money, advertis, will, gold
FREX: contradict, money, profit, bet, gold, buy, advertis
Lift: anti-x, dubya, monit, pepsi, pro-x, upkeep, ess
Score: site, money, reddit, advertis, gold, profit, contradict

Topic 46 Top Words:
Highest Prob: line, argument, draw, societi, continu, racist, ratheism
FREX: ratheism, line, draw, nazi, genocid, emot, rchristian
Lift: craze, invers, man-children, <URL>, repar, unapologet, unmanag
Score: line, ratheism, draw, argument, nazi, societi, belief

Topic 47 Top Words:
Highest Prob: sure, pretti, much, racism, caus, sound, can
FREX: pretti, sure, sound, slope, slipperi, sexism, safeti
Lift: deafen, ditto, endeavor, hate-centr, institutionalingrain, nauseat, pond
Score: pretti, sure, racism, sound, slope, slipperi, logic

Topic 48 Top Words:
Highest Prob: mod, user, remov, moder, admin, can, sub
FREX: mod, delet, remov, moder, report, user, account
Lift: banter, cetera, flub, <URL>, modqueu, oneself, relinquish
Score: mod, user, moder, remov, delet, admin, account

Topic 49 Top Words:
Highest Prob: now, chang, specif, year, pleas, two, help
FREX: pleas, hes, specif, now, ago, chang, wait
Lift: flip-flop, flop, <URL>, turtl, bingo, rcrackertown, dworkin
Score: specif, now, chang, year, pleas, hes, ago

Topic 50 Top Words:
Highest Prob: get, reddit, can, mean, complet, someth, isnt
FREX: complet, get, mean, general, someth, deal, lot
Lift: mysoginist, discont, scholarship, eras, complet, disabl, general
Score: get, complet, reddit, mean, general, someth, can

---

## Topic list for 2018 Content Policy:

---

Topic 1 Top Words:
Highest Prob: dont, say, theyr, need, anyth, claim, liber
FREX: theyr, extremist, dont, safe, eye, refut, wors
Lift: neglect, quash, rob, bob, favourit, -touch, aight
Score: dont, say, theyr, need, liber, violat, safe

Topic 2 Top Words:
Highest Prob: subreddit, rule, someth, ask, that, respons, guy
FREX: subreddit, rall, joke, rpopular, humor, rule, ask
Lift: rebuttel, rpopular, subredit, anyhow, balloon, bark, blackout
Score: subreddit, rule, rall, break, joke, moder, ask

Topic 3 Top Words:
Highest Prob: ban, speech, hate, free, read, first, exact
FREX: voat, hate, echochamb, ban, hill, speech, infest

Lift: brightest, durr, lib, alan, articul, banning", blackpeopletwitt

Score: ban, speech, hate, free, read, voat, first

Topic 4 Top Words:

Highest Prob: post, comment, look, link, see, still, mod

FREX: link, post, delet, rthedonald, comment, repli, front

Lift: "genocid, anti-semet, atleast, baizous, boner, ceddit, cherry-pick

Score: post, comment, link, delet, page, mod, downvot

Topic 5 Top Words:

Highest Prob: left, trump, vote, statement, elect, presid, republican

FREX: republican, hillari, berni, candid, fox, kavanaugh, clinton

Lift: benghazi, counti, creepi, disservic, fragility", gateway, hmmmm

Score: trump, hillari, vote, left, republican, clinton, kavanaugh

Topic 6 Top Words:

Highest Prob: world, countri, live, state, capit, capitalist, social

FREX: institut, slaveri, modern, initi, usa, capitalist, educ

Lift: institut, anti-vaxx, await, british, circus, coloni, conflict-driven

Score: countri, capit, capitalist, world, educ, america, modern

Topic 7 Top Words:

Highest Prob: nazi, kill, genocid, fascist, holocaust, peopl, million

FREX: genocid, nazi, fascist, mao, palestinian, holocaust, fascism

Lift: albeism, arabia, axi, bastard, ceausescu, ceau\u0219escu, cth

Score: nazi, genocid, fascist, kill, famin, ethnic, jew

Topic 8 Top Words:

Highest Prob: much, platform, can, privat, social, protect, compani

FREX: compani, gold, protect, privat, platform, slope, pressur

Lift: alter, bakeri, chariti, confront, freez, fuckwit, guardian

Score: privat, compani, platform, social, protect, govern, media

Topic 9 Top Words:

Highest Prob: peopl, make, one, actual, said, never, happen

FREX: make, money, thought, 're, though, fun, happen

Lift: blackjack, dementia, facebookinstagramtwitt, fascin, feelingsens, rsanctionedsuicid, vial

Score: peopl, make, 're, never, talk, money, actual

Topic 10 Top Words:

Highest Prob: exampl, year, problem, side, either, issu, brigad

FREX: aaron, exampl, harass, insult, brigad, last, late

Lift: againsthatesubreddit, constitu, constrain, euro, orient, seeth, streak

Score: exampl, brigad, year, side, harass, insult, last

Topic 11 Top Words:

Highest Prob: reddit, will, admin, see, even, sure, keep

FREX: reddit, thank, admin, keep, will, continu, next

Lift: rcanada, rstopadvertis, ad-friend, afloat, apocalyps, beard, bogeyman

Score: reddit, admin, will, thank, keep, step, sure

Topic 12 Top Words:

Highest Prob: polit, violenc, open, rpolit, ignor, chamber, echo

FREX: chamber, echo, poster, rpolit, violenc, rtd, open

Lift: abhor, ahso, anti-lgbt, aswel, <URL>, confisc, corps

Score: violenc, rpolit, polit, echo, chamber, open, ideolog

Topic 13 Top Words:

Highest Prob: fuck, shit, time, liter, everi, get, yeah

FREX: god, lsc, ass, shit, fuck, bullshit, yeah

Lift: villag, abet, accid, aggreanc, ala, bingo, cellshttparchiveisvpax

Score: fuck, shit, liter, everi, time, bullshit, god

Topic 14 Top Words:

Highest Prob: right, know, call, mean, doesnt, just, propaganda

FREX: russian, propaganda, wing, troll, faggot, victim, spread

Lift: wink, ayup, lick, boat, breadlin, castil, clear-cut

Score: right, propaganda, call, russian, spread, wing, mean

Topic 15 Top Words:
    Highest Prob: thing, censorship, site, view, censor, allow, seem
    FREX: desktop, access, censor, site, mobil, censorship, toxic
    Lift: comcast, applaus, benjamin, bowti, bumpi, duck, foul
    Score: censorship, mobil, site, censor, access, view, websit

Topic 16 Top Words:
    Highest Prob: want, white, idea, peopl, black, major, okay
    FREX: fragil, jone, skin, okay, white, want, idea
    Lift: anyone', fee, latino, mayocid, psyop, -group, alley
    Score: white, want, black, idea, oppress, okay, race

Topic 17 Top Words:
    Highest Prob: get, support, trump, word, stupid, man, pleas
    FREX: retard, nigga, orang, mate, dick, rid, lmao
    Lift: armenian, automatiskt, beta, bless, bold, dairi, denna
    Score: trump, support, get, stupid, pleas, retard, man

Topic 18 Top Words:
    Highest Prob: good, cant, tri, lol, take, got, arent
    FREX: yike, lol, got, keto, good, arent, guess
    Lift: calori, inabl, interbre, mutil, "chines, botshill", bow
    Score: lol, good, got, tri, cant, arent, big

Topic 19 Top Words:
    Highest Prob: case, group, great, also, even, term, job
    FREX: case, fine, owner, particular, basi, impli, refer
    Lift: cancel, jeong, kotkin, valeri, ann, coulter, enoch
    Score: case, slave, term, group, great, standard, basi

Topic 20 Top Words:
    Highest Prob: fact, believ, agre, power, definit, must, accept
    FREX: must, fact, truth, agre, truli, equal, valu
    Lift: abnorm, dint, falsehood, hypothesi, interrog, orwel, overus
    Score: believ, agre, fact, power, truth, must, definit

Topic 21 Top Words:
    Highest Prob: think, just, your, realli, bad, can, someon
    FREX: your, bad, realli, tldr, think, defend, sad
    Lift: cringey, gallop, gish, glorious, goup, helperbot, lone
    Score: your, think, opinion, bad, realli, someon, just

Topic 22 Top Words:
    Highest Prob: like, use, well, racist, don't, stop, racism
    FREX: shill, misogyni, racist, block, poseidon, like, spacedick
    Lift: alright, bluewav, bookmark, ftfi, jimmi, poseidon, shill
    Score: like, don't, racist, use, well, stop, racism

Topic 23 Top Words:
    Highest Prob: sub, quarantin, can, communiti, user, content, list
    FREX: quarantin, nsfw, accident, revenu, sub, heads-, search
    Lift: "ban", aboutjson, accident, accross, blank, canari, commentspost
    Score: quarantin, sub, content, user, communiti, offens, list

Topic 24 Top Words:
    Highest Prob: now, way, point, differ, come, let, chang
    FREX: feminist, femin, now, women, turn, men, male
    Lift: anti-tran, cannib, condescend, desies, femin, globalist, honey
    Score: now, way, men, chang, let, come, point

Topic 25 Top Words:
    Highest Prob: everyon, els, peopl, like, way, know, say
    FREX: everyon, els, consequ, mine, arm, daughter, depend
    Lift: antagon, cobra, cousin, hanniti, joo, milleni, oklahoma
    Score: everyon, els, opium, streisand, peopl, fuck, gofundm

Topics of 2020 Content Policy update:

Topic 1 Top Words:
    Highest Prob: actual, guy, lmao, absolut, shut, ill, ass
    FREX: shut, rbigchungus, lmao, trash, asshol, holi, soyboy
    Lift: boooooooo, cop-, groundwork, jerkofftomymom, latt, moli, oppression"
    Score: guy, actual, lmao, absolut, shut, ass, asshol
Topic 2 Top Words:
    Highest Prob: come, give, work, back, money, fight, kid
    FREX: digg, give, asylum, come, money, fund, ton
    Lift: alexa, alrighti, argentium, de-human, defundreddit, disastr, fark
    Score: give, back, come, money, work, award, illeg
Topic 3 Top Words:
    Highest Prob: never, ive, ever, man, person, seen, life
    FREX: ive, never, hey, seen, dumbest, ever, listen
    Lift: drip-f, <URL>, kneejerk, rapper, self-respect, sheep", statistics"
    Score: never, ive, seen, man, ever, hey, life
Topic 4 Top Words:
    Highest Prob: reddit, fuck, place, websit, bullshit, leav, done
    FREX: reddit, websit, leav, place, fuck, yea, bullshit
    Lift: kristallnacht, mfw, party', puta, rexxit, <URL>, assho
    Score: reddit, fuck, websit, place, leav, bullshit, done
Topic 5 Top Words:
    Highest Prob: great, squar, tiananmen, forward, 李洪志, 法輪功, tibet
    FREX: squar, tiananmen, 李洪志, 法輪功, tibet, 天安門, 天安门
    Lift: tibet, abduct, anti-protect, anti-rightist, anti-riot, baggag, epoch
    Score: 李洪志, 法輪功, 天安門, 天安门, great, falun, tiananmen
Topic 6 Top Words:
    Highest Prob: year, activ, month, ago, basic, wont, wait
    FREX: didn't, yall, ago, gamer, year, rwojak, month
    Lift: rhongkong, rtrump, aha, cofound, gamer, hategroup, <URL>
    Score: year, ago, month, rsino, donald, wait, didn't
Topic 7 Top Words:
    Highest Prob: china, countri, system, parti, cultur, america, democrat
    FREX: ccp, cancel, own, democraci, tencent, parti, econom
    Lift: conspiratori, militar, anti-asian, asset, ccp, cpc, denmark
    Score: china, democrat, system, countri, america, cultur, parti
Topic 8 Top Words:
    Highest Prob: even, though, stop, pretend, doubt, chanc, unlik
    FREX: someone', uundeletepar, <URL>, ripreddit, <URL>, <URL>, doubt
    Lift: chunga, gottem, indubit, les, lmaoooooo, ripreddit, someone'
    Score: even, though, stop, pretend, doubt, unlik, chanc
Topic 9 Top Words:
    Highest Prob: realiti, behavior, soon, repeat, essenti, ahead, dare
    FREX: soon, hello, touch, horseshit, mask, essenti, incorrect
    Lift: bane, blockad, crucial, fidel, tienanmen, tumor, wahhhh
    Score: realiti, behavior, soon, repeat, touch, dare, ahead
Topic 10 Top Words:
    Highest Prob: see, happen, side, love, god, stand, hear

FREX: side, hear, bless, love, lefti, god, happen
Lift: amen, cumrad, bless, leisur, seibel, semper, side
Score: see, side, happen, love, god, hear, stand

Topic 11 Top Words:
    Highest Prob: women, men, tran, male, space, feminist, safe
    FREX: exclusionari, men, feminist, women, misogynist, safe, misogyni
    Lift: "biolog, bachelorbachelorett, commiser, gendercriticalguy, inventori, lipstick, mansplain
    Score: women, men, tran, feminist, male, space, femin

Topic 12 Top Words:
    Highest Prob: ban, sub, subreddit, got, chapo, slave, cth
    FREX: subreddit, sub, ban, evas, got, mad, chapo
    Lift: campshttpsiimgurcomwetfjpjpg, deadlyfetish, lad, otoh, surreal, thediddl, trans-friend
    Score: ban, sub, subreddit, chapo, got, slave, owner

Topic 13 Top Words:
    Highest Prob: will, now, sinc, open, appar, rest, longer
    FREX: shitpost, blah, offici, now, will, pack, sinc
    Lift: 🧑, aye, dab, devolut, downstream, fullscreen, <URL>
    Score: now, will, open, sinc, offici, appar, rest

Topic 14 Top Words:
    Highest Prob: much, sure, pretti, obvious, lie, full, funni
    FREX: pcm, cumtown, redditor, fragil, fbi, satir, sure
    Lift: banter, boooooo, goat, hmmmm, moulderin, offmychest, pcm
    Score: sure, pretti, lie, much, fragil, obvious, troll

Topic 15 Top Words:
    Highest Prob: still, vote, etc, repli, popular, rgendercrit, worri
    FREX: rchodi, worri, strang, rgendercrit, misogynyfetish, hillari, rcommun
    Lift: deadey, kurd, oust, pinch, rabuseporn, rbanfemalehatesub, rbanmalehatesub
    Score: still, vote, repli, rgendercrit, popular, worri, bet

Topic 16 Top Words:
    Highest Prob: talk, rape, porn, argu, consent, video, child
    FREX: kink, consensu, porn, rape, fantasi, fetish, talk
    Lift: dimorph, gentl, hentai, irrit, ampute, charm, cnc
    Score: rape, porn, talk, consent, kink, fantasi, child

Topic 17 Top Words:
    Highest Prob: polit, left, conserv, liber, far, leftist, ideolog
    FREX: lean, left, polit, agenda, left-w, liber, leftist
    Lift: lean, swallow, adguardhttpadguardcom, bla, boon, breather, chrome
    Score: left, polit, leftist, liber, conserv, chamber, echo

Topic 18 Top Words:
    Highest Prob: can, anyon, tell, find, pleas, els, somewher
    FREX: somewher, anyon, tell, can, find, els, pleas
    Lift: 🎃 🎃 🎃 🎃 🎃 🎃 🎃 🎃, emir, <URL>, mysoginist, oman, oprah
    Score: can, anyon, pleas, tell, find, els, somewher

Topic 19 Top Words:
    Highest Prob: good, bad, clear, alreadi, must, faith, final
    FREX: faith, bad, alreadi, reddit—, good, clear, riddanc
    Lift: "view, anti-corpor, bad, groomer, pickl, reddit—, shoo
    Score: good, bad, faith, clear, alreadi, final, spectrum

Topic 20 Top Words:
    Highest Prob: post, comment, remov, made, account, edit, thread
    FREX: post, comment, report, account, button, link, remov
    Lift: tuesday, hyperlink, <URL>, threshold, -day, alliter
    Score: post, comment, report, account, thread, link, remov

Topic 21 Top Words:
    Highest Prob: real, gone, best, learn, told, voic, pick
    FREX: gone, bootlick, voic, folk, left-lean, learn, 🎃 🎃 🎃 🎃 🎃

Lift: parad, blaze, calvin, diversifi, foreskin, gone, hairi

Score: gone, real, learn, best, voic, pick, told

Topic 22 Top Words:

Highest Prob: mod, user, communiti, power, action, taken, rthedonald

FREX: rthedonald, mod, mute, inact, communiti, awe, trip

Lift: pigpoopballsjpg, rireland, spook, stinki, это, 🎃🎃🎃, 🎃🎃🎃🎃🎃🎃

Score: mod, user, communiti, rthedonald, team, power, abus

Topic 23 Top Words:

Highest Prob: support, yeah, trump, nazi, evid, today, easi

FREX: bro, doj, yeah, booster, <URL>, counterproduct

Lift: "fair, "watchredditdie", 🎃🎃🎃shut, autogynephilia, awww, bernard, blanchard

Score: trump, nazi, yeah, support, evid, beauti, bro

Topic 24 Top Words:

Highest Prob: realli, understand, ask, ignor, serious, respons, honest

FREX: understand, realli, ask, serious, went, ignor, interest

Lift: pseudo, bark, frail, urgh, went, understand, curios

Score: realli, understand, ask, ignor, serious, honest, interest

Topic 25 Top Words:

Highest Prob: your, wrong, noth, there, critic, dude, what

FREX: cunt, your, wrong, noth, karen, dumpster, truli

Lift: apach, cht, tramra, cunt, dumpster, felicia, zoom

Score: your, wrong, noth, dude, what, critic, youv

Topic 26 Top Words:

Highest Prob: shit, thank, uspez, question, least, answer, dumb

FREX: unwordcountbot, dumb, liar, answer, question, uspez, thank

Lift: 👏👏👏, archiveorg, rgaymarriag, shove", trashi, yorker, "pleas

Score: shit, thank, uspez, question, answer, piec, dumb

Topic 27 Top Words:

Highest Prob: rule, content, new, admin, polici, moder, break

FREX: enforc, content, break, moder, broke, new, violat

Lift: "nuh, "violent, content", defianc, instaban, non-nest, pikabu

Score: rule, content, admin, new, moder, polici, break

Topic 28 Top Words:

Highest Prob: don't, 're, definit, 've, doesn't, can't, isn't

FREX: don't, can't, isn't, aren't, 're, doesn't, haven't

Lift: "just, populist, rehabilit, "certain, "expand, action", aren't

Score: 're, don't, definit, 've, doesn't, isn't, can't

Topic 29 Top Words:

Highest Prob: right, wing, stupid, realiz, most, lgbt, none

FREX: stupid, wing, right, kinda, lul, lgbt, realiz

Lift: auth-centr, ditto, dropthet, <URL>, neoliberalist, persuas, privatis

Score: right, wing, stupid, realiz, lgbt, kinda, extremist

Topic 30 Top Words:

Highest Prob: keep, gonna, speak, cri, respect, pay, worth

FREX: gonna, fan, boot, keep, speach, speak, observ

Lift: angrier, ccps, chuckl, mma, quicker, recov, richard

Score: keep, gonna, cri, speak, pay, figur, eye

Topic 31 Top Words:

Highest Prob: suck, wow, moral, statement, evil, dick, boy

FREX: suck, cock, brave, authent, wow, lectur, moral

Lift: "blm', "deplatform", aros, booo, brave, cock, comed

Score: suck, wow, dick, evil, moral, statement, couldnt

Topic 32 Top Words:

Highest Prob: posit, mental, brain, effect, chang, articl, scienc

FREX: dysphoria, consensus, scientif, studi, indic, health, brain

Lift: gender-affirm, genealog, academi, accredit, adderal, anthropologist, anti-depress

Score: transit, mental, studi, brain, scienc, posit, dysphoria

Topic 33 Top Words:
Highest Prob: think, mean, that, doesnt, cant, isnt, theyr
FREX: sad, yup, doesnt, that, mean, weird, rfuckyoukaren
Lift: hahahahahahaha, repar, rfuckyoukaren, yup, anthropomorph, breakfast, inkl
Score: that, mean, doesnt, think, theyr, cant, yes

Topic 34 Top Words:
Highest Prob: allow, base, certain, human, rememb, skin, color
FREX: rblackpeopletwitt, send, verifi, bpt, rwhitepeopletwitt, april, forearm
Lift: bpt, verifi, black-, check-mark, non-black, nonwhite-, rblackworldord
Score: skin, allow, rblackpeopletwitt, color, rememb, club, verifi

Topic 35 Top Words:
Highest Prob: kill, polic, die, death, murder, genocid, communist
FREX: tanki, islam, holocaust, janni, communism, polic, iirc
Lift: holocaust, janni, lenin, patrol, strongest, unbridl, "religion
Score: kill, polic, genocid, cop, murder, communist, death

Topic 36 Top Words:
Highest Prob: speech, free, freedom, privat, law, allow, govern
FREX: amend, privat, speech, free, law, freedom, servic
Lift: amend, government, megacorpor, non-fre, opress, pepperidg, roam
Score: speech, free, freedom, privat, law, amend, compani

Topic 37 Top Words:
Highest Prob: major, group, protect, rule, hate, minor, promot
FREX: group, minor, major, protect, promot, global, margin
Lift: "disenfranchised", "hangfs", "killallnrs", "major, "protect, disability", hate"
Score: group, major, protect, rule, ident, hate, minor

Topic 38 Top Words:
Highest Prob: believ, part, sorri, miss, christian, due, level
FREX: haha, buddi, impli, bibl, sorri, believ, ironi
Lift: buddi, delici, funniest, autogynephil, breathtak, crucifi, deprec
Score: believ, sorri, part, christian, miss, jesus, level

Topic 39 Top Words:
Highest Prob: alway, fascist, hatr, societi, fascism, win, toler
FREX: ruin, fascism, scam, win, lurk, paradox, hatr
Lift: cutur, extinct, flippant, holotyp, missionari, mussolini, playbook
Score: alway, hatr, fascist, fascism, societi, win, toler

Topic 40 Top Words:
Highest Prob: gender, tran, sex, woman, terf, femal, peopl
FREX: transphob, lesbian, vagina, genit, terf, feminin, transphobia
Lift: feminin, gnc, monosexu, phobic, prefix, testicl, underdevelop
Score: gender, sex, tran, femal, woman, terf, lesbian

Topic 41 Top Words:
Highest Prob: site, becom, run, move, mind, enjoy, ruqqus
FREX: site, ruqqus, chan, voat, migrat, server, app
Lift: backer, bitchut, catch-, dent, discordhttpsdiscordggnday, donald', humanitarian
Score: site, ruqqus, move, becom, join, run, chan

Topic 42 Top Words:
Highest Prob: big, decid, continu, quit, compani, corpor, busi
FREX: wtf, tech, big, quit, behind, chungus, silicon
Lift: baker, glorious, blackshirt, bust, chungus, chyna, eco
Score: big, compani, decid, quit, continu, corpor, busi

Topic 43 Top Words:
Highest Prob: say, racist, call, okay, exact, world, fine
FREX: doubl, subtl, indian, fine, racist, fwr, say
Lift: "asian, "rich, azn, chapotard, gambit, rindianpeoplefacebook, subt
Score: racist, say, okay, asian, exact, fine, call

Topic 44 Top Words:
    Highest Prob: violenc, rchapotraphous, call, deserv, chapotraphous, user, celebr
    FREX: chapotraphous, uniron, celebr, rchapotraphous, violenc, hat, deserv
    Lift: boys<URL>, charl, <URL>
    Score: rchapotraphous, violenc, chapotraphous, amidst, congressmen, defranco, kidsthes
Topic 45 Top Words:
    Highest Prob: state, general, bit, number, half, unit, hitler
    FREX: intellig, draw, hitler, hail, hammer, bit, general
    Lift: iranian, magnitud, plate, prosper, protein, quantum,
    Score: state, bit, number, unit, half, hitler, general
Topic 46 Top Words:
    Highest Prob: peopl, white, black, racism, race, person, discrimin
    FREX: racism, redefin, white, identitarian, supremaci, peopl, race
    Lift: -hate, 'white, 'whiteness', "institut, avarag, caucasus, cherokeeirishgerman
    Score: white, peopl, black, racism, race, discrimin, supremacist
Topic 47 Top Words:
    Highest Prob: everi, day, start, yet, old, singl, week
    FREX: everi, slipperi, old, comedi, yet, week, singl
    Lift: bailli, comedi, oceania, "liber, and, assail, carreer
    Score: everi, day, yet, old, singl, start, week
Topic 48 Top Words:
    Highest Prob: need, read, next, write, situat, transpar, book
    FREX: congrat, kill, laden, transpar, praxi, comprehens, read
    Lift: amber, beleiv, bottl, congrat, kobe, praxi, rspez
    Score: need, read, next, write, transpar, situat, kill
Topic 49 Top Words:
    Highest Prob: tri, someon, reason, better, issu, either, less
    FREX: tri, reason, better, explain, hard, either, instead
    Lift: "whites", rjustbblack, rjustbewhit, doubli, "feminist", blast, melodramat
    Score: tri, someon, better, reason, explain, instead, less
Topic 50 Top Words:
    Highest Prob: make, chines, put, sens, especi, propaganda, truth
    FREX: rchina, cring, chines, hmm, make, sens, appreci
    Lift: dilat, non-us, cring, egyptian, flew, hmm, platinum
    Score: make, chines, put, sens, propaganda, western, rchina
Topic 51 Top Words:
    Highest Prob: hope, republican, fuck, peopl, like, make, die
    FREX: patriot, tree, gop, hope, republican, metro, quiet
    Lift: asswip, flyover, <URL>, <URL>, <URL>, <URL>, <URL>
    Score: republican, hope, fuck, tree, agnost, already-occur, arsenic
Topic 52 Top Words:
    Highest Prob: take, away, cours, step, notic, properti, wouldn't
    FREX: "'m, advic, sharehold, step, unjust, notic, rdankmem
    Lift: "'m, consolid, cyber, elections", embargo, evict, farcic
    Score: take, away, cours, properti, step, notic, wouldn't
Topic 53 Top Words:
    Highest Prob: get, anyth, spez, joke, littl, downvot, imagin
    FREX: spez, downvot, pathet, nice, surpris, neckbeard, <URL>
    Lift: erot, non-leftist, umanan, 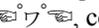, coron, de-upvot, ember
    Score: get, spez, anyth, joke, downvot, upvot, imagin
Topic 54 Top Words:
    Highest Prob: time, mani, long, first, page, sort, saw
    FREX: loop, hatespeech, long, time, theyv, mani, multipl
    Lift: biscuit, ccp's, conspiracist, eta, <URL>, internet", login
    Score: time, mani, long, first, page, saw, front
Topic 55 Top Words:

Highest Prob: just, like, dont, look, someth, know, seem
FREX: like, just, seem, look, dont, sound, someth
Lift: voice", sext, limp, <URL>-<URL>, obama', sheesh, sound
Score: like, just, dont, look, seem, someth, sound

Topic 56 Top Words:
Highest Prob: thing, mayb, help, around, stuff, rather, amount
FREX: mayb, welcom, crazi, help, forget, thing, blm
Lift: amigo, rjohnbrown, welcom, crazi, anti-imperialist, forget, mlk
Score: thing, mayb, help, stuff, around, welcom, blm

Topic 57 Top Words:
Highest Prob: feel, let, problem, kind, harass, line, target
FREX: problem, suppos, hit, kind, sit, line, shitti
Lift: <URL>, "liberals", bleep-bloop
Score: feel, problem, let, harass, hurt, kind, target

Topic 58 Top Words:
Highest Prob: end, news, age, normal, begin, remain, hundr
FREX: purg, wrongthink, blow, nazbol, age, kek, cent
Lift: "supremacy", ame, asphyxi, batman, black-lives-matt, boogi, deposit
Score: end, news, age, begin, purg, normal, wrongthink

Topic 59 Top Words:
Highest Prob: caus, fair, turn, twitter, stay, defend, hand
FREX: nah, fair, pander, stay, fee, twitter, caus
Lift: altright, adulthood, anti-fasc, covet, demonet, dragon, fahrenheit
Score: caus, twitter, fair, stay, turn, defend, nah

Topic 60 Top Words:
Highest Prob: last, check, sourc, attempt, wonder, privileg, bot
FREX: usernam, itll, cum, check, sourc, contact, last
Lift: femboy, sooooo, 'privelidge', "get, beep, clubhous, cum
Score: check, last, sourc, bot, privileg, wonder, usernam

Topic 61 Top Words:
Highest Prob: idea, platform, censorship, social, internet, media, control
FREX: censorship, internet, discours, liabl, forum, idea, media
Lift: electron, fcc, scarier, watchpeopledi, weakl, yishan, "cause"
Score: platform, censorship, media, idea, social, internet, forum

Topic 62 Top Words:
Highest Prob: way, everyon, without, equal, view, case, job
FREX: way, everyon, equal, hold, goe, abl, without
Lift: bandaid, insul, whiteblack, brexit, naiveti, neo-marxist, poof
Score: way, everyon, equal, without, view, job, abl

Topic 63 Top Words:
Highest Prob: point, opinion, differ, agre, disagre, other, whole
FREX: opinion, agre, disagre, point, silenc, whole, happi
Lift: "silenc, bum, terfs", verbatim, pregaru, opinion, wallet
Score: opinion, agre, point, differ, disagre, silenc, whole

Topic 64 Top Words:
Highest Prob: thought, elect, valu, capit, insult, forc, capitalist
FREX: insult, wage, capit, capitalist, dissent, enemi, profit
Lift: "slaveowners", bike, counteract, dau, entrepreneur, festiv, hurdl
Score: thought, capit, elect, capitalist, insult, valu, worker

Topic 65 Top Words:
Highest Prob: liter, lol, hes, presid, fail, trigger, lib
FREX: lib, loser, forev, rpoliticalcompassmem, salti, hog, rot
Lift: aaand, anti-right, antirac, armband, berniebro, dork, drinker
Score: lol, liter, lib, hes, bitch, trigger, presid

Topic 66 Top Words:
Highest Prob: care, wouldnt, bigot, friend, anyway, board, game

FREX: care, friend, humor, board, cheer, token, bigot
Lift: mmmm, tasti, token, humor, joel, pineappl, prn
Score: care, bigot, friend, board, game, wouldnt, deal

Topic 67 Top Words:
Highest Prob: didnt, fact, enough, american, wasnt, anti, follow
FREX: didnt, anti, rconsumeproduct, wasnt, consumer, enough, gave
Lift: rconsumeproduct, anti, anti-comsumer, anti-sex, anticonsumer, didnt, <URL>
Score: didnt, fact, american, enough, wasnt, anti, recent

Topic 68 Top Words:
Highest Prob: one, anoth, entir, name, list, two, top
FREX: fella, rfragileblackredditor, rsmuggi, top, forgot, sir, one
Lift: coerceblackmailguiltrip, exept, freakout, <URL> , rclericalfasc, rsoyboy, tamer
Score: one, list, name, top, anoth, two, heard

Topic 69 Top Words:
Highest Prob: hate, use, word, consid, direct, accept, defin
FREX: word, english, hate, direct, use, consid, defin
Lift: mandarin, rcc, "huger", banharass, biblic, male-h, raskpinkpil
Score: hate, use, word, defin, consid, english, languag

Topic 70 Top Words:
Highest Prob: exist, probabl, type, nobodi, total, insan, sometim
FREX: yep, antifa, smart, probabl, insan, nobodi, fake
Lift: <URL> , rblackpow, sankara, smart, trans-support, weirdo, yep
Score: probabl, exist, nobodi, type, total, insan, antifa

Topic 71 Top Words:
Highest Prob: also, well, said, guess, claim, true, might
FREX: guess, well, said, rid, idiot, true, complet
Lift: flaiano, foil, monk, cheek, guess, <URL> , rid
Score: said, well, guess, true, also, might, complet

Topic 72 Top Words:
Highest Prob: meme, correct, share, context, refer, civil, similar
FREX: meme, whistl, boogaloo, share, boog, throw, context
Lift: anxious, atf, bird, electr, herp, pupper, stochast
Score: meme, correct, share, civil, war, context, refer

Topic 73 Top Words:
Highest Prob: face, theori, near, along, conspiraci, reach, sentenc
FREX: reach, libertarian, ain't, cartoon, horsesho, inaccur, tactic
Lift: "look, ammo, beef, buncha, carlin, cartoon, cousin
Score: theori, near, along, libertarian, conspiraci, reach, face

Topic 74 Top Words:
Highest Prob: live, matter, slaveri, black, famili, action, middl
FREX: slaveri, matter, live, centuri, irish, east, middl
Lift: "whiteness", appalachia, bacteria, boston, boyz, den, eve
Score: live, matter, slaveri, black, middl, poverti, africa

Topic 75 Top Words:
Highest Prob: want, know, think, just, also, chang, much
FREX: want, know, think, also, just, chang, much
Lift: want, know, think, also, chang, just, much
Score: want, know, think, just, also, chang, much